# Automatic driving path plan based on iterative and triple optimization method

Yang Yinyang[1]   Wang Chanchan[2]

1. Beijing JY Intelligent Technology Co. Ltd BeiJing 101300;
2. Beijing Institute of Technology BeiJing 101300;

**Abstract: This paper presents a triple optimization algorithm of two-dimensional space, driving path and driving speed, and iterates in the time dimension to obtain the local optimal solution of path and speed in the optimal driving area. Design iterative algorithm to solve the best driving path and speed within the limited conditions. The algorithm can meet the path planning needs of automatic driving vehicle in complex scenes and medium and high-speed scenes.**

## Introduction

Local trajectories planning of autonomous vehicles connected with localization, perception and prediction. Generate a collision free and comfortable path according to localization information, environment and map elements [1][6]. The performance of trajectories planning algorithm determines the safety, driving flexibility and comfort of autonomous vehicles. In this paper, the optimal convex space combination is obtained by splitting, reorganizing and optimizing the nonconvex space. In the convex space combination, the quadratic programming method is used to optimize the path. Autonomous vehicles make decisions on obstacles and generate convex space in the range of 's-t'. The quadratic programming method is used to optimize the speed. The optimal driving path and speed under the restricted conditions are realized by iterative solution.

## 1. Triple optimization algorithm design

Automatic driving path planning algorithm needs to carry out trajectory and speed planning in two-dimensional space and time dimension in a short time. Currently, the common ideas of path planning include search based[2], optimization based on batch path sampling[3], combination based on search and optimization calculation[4], and optimization calculation. The common ideas of speed planning include optimization based on batch sampling, combination based on search and optimization calculation [3], and optimization calculation. Each approach has its own applicable scenarios. The difficulty of the method based on optimization is the generation of convex space and the optimization of computation quantity.

However, as a planning algorithm with clear process and strong applicability, it has obvious advantages in wide application scenarios and dynamic obstacle handling.

### 1.1 Convex space optimization

In the process of driving, other vehicles and pedestrians in the environment change the driving area into a non-convex area [5], as shown in Figure 1.

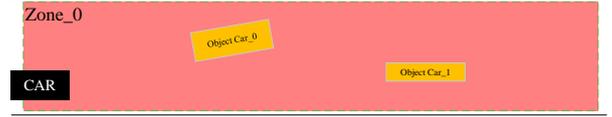

Figure 1 Driving area is non-convex space

According to the position of obstacles in the *frenet* coordinate system, the original non-convex area is partitioned and reassembled. The original non-convex space Zone_0 is divided into Zone_0_0, Zone_1_0, Zone_1_1, Zone_2_0, Zone_3_0 Zone_3_1 and Zone_4_0, as shown in Figure 2.

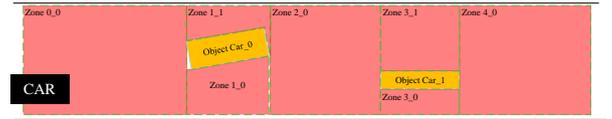

Figure 2 Split non-convex space

After splitting the convex set, define the elements in each convex set to include the following geometric attributes:

*parent pointer*: the pointer points to the address of the previous convex space.

*area cost*: the ratio of the area of the convex space to the maximum area of the convex space.

*link cost*: subspace is related to parent space link length.

*lane cost*: convex space contains the area of the destination lane related.

According to the geometric loss function, to find the optimal convex combination in the convex set, ergodic, A* and other algorithms can be adopted [1].

However, if only consider the geometric loss of convex set, the dynamic characteristics of future driving cannot be well controlled, so the decision tree is built according to the geometric relationship of convex set. The decision Tree is shown in Figure 3.

As shown in Figure 3, the decision tree demonstrates possible decision behaviors in a vehicle. Each path of the decision tree (which may not reach the leaf node) corresponds to a vehicle decision behavior.

For each path of the decision tree, referred to as a convex combination, the defined loss is as follows:

*MoveCost*: The ratio of the length of the combination to the maximum length of all combinations.

*PathCost*: Future steering wheel angle.

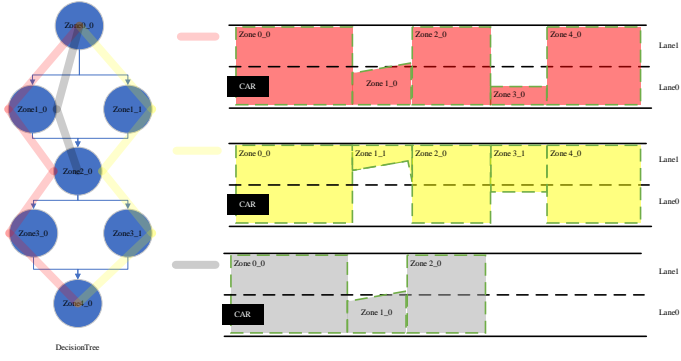

Figure 3 Decision-Tree

***DynamicCost***: Future lateral and longitudinal acceleration.

After generating the decision tree, in scenarios where there are numerous branches and an abundance of decision options, this paper proposes that before calculating ***PathCost*** and ***DynamicCost***, the geometric losses (sum ***area_cost*** and ***MoveCost***) of a set of decisions should be roughly sorted, and the ***20*** decisions with the smallest losses should be selected for subsequent calculations

***DynamicCost*** and ***PathCost*** represent considerations of vehicle smoothness in future driving when making current decisions. This paper uses acceleration and steering wheel angle change (replaced by path curvature) to characterize these losses.

For a given decision, coarse sampling is conducted in the ***s*** direction and sampled point constraints are generated.

Sampled point constraints are listed as follows:

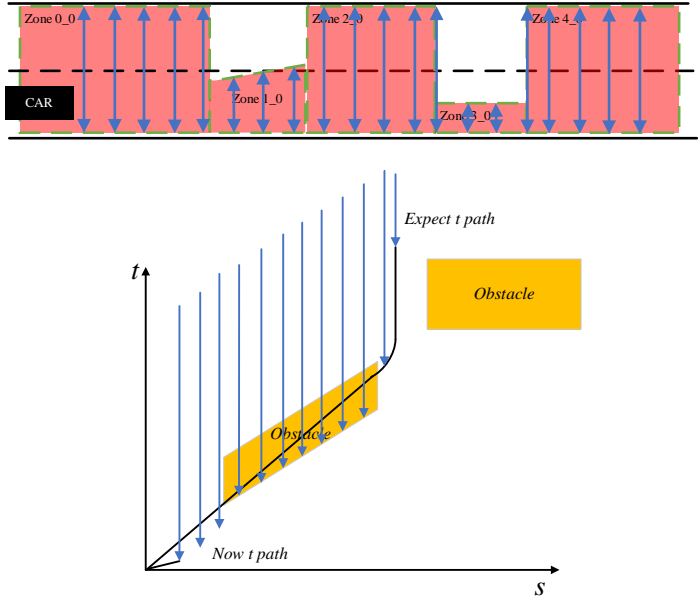

Figure 3 Sample boundary generation

During the sampling process of a decision, constraints for different sampling positions are generated. In the sampling process for lateral acceleration, a ***t-s*** graph is utilized to generate the lower boundary for time ***t***.

After obtaining the coarse sampling under a decision, a quadratic programming approach is used to compute the coarse trajectory, which in turn provides information on the curvature of future driving, as well as the variations in longitudinal acceleration and lateral acceleration.

Optimization variables:

$$l_0 \quad l_1 \quad l_2 \quad ... \quad l_{i-1} \quad l_i \quad l_{i+1} \quad t_0 \quad t_1 \quad t_2 \quad ... \quad t_{i-1} \quad t_i \quad t_{i+1} \tag{1}$$

$l_i, t_i$ Shows the lateral coordinate of the sampling point in the road coordinate system and sampling time.

Target：

$$\cos t = \sum_1^n (l_{i-1} + l_{i+1} - 2l_i)^2 + \sum_0^n (l_i - l_{ref})^2 + \sum_1^n (t_{i-1} + t_{i+1} - 2t_i)^2 + \sum_0^n (t_i - t_{lb})^2 \tag{2}$$

$\sum_1^n (l_{i-1} + l_{i+1} - 2l_i)^2$ Smoothness metrics for sampled trajectories.

$\sum_0^n (l_i - l_{ref})^2$ Deviation of sampled trajectories from the centerline.

$\sum_1^n (t_{i-1} + t_{i+1} - 2t_i)^2$ Acceleration and deceleration smoothness metrics for sampled trajectories.

$\sum_0^n (t_i - t_{lb})^2$ Expresses the rapid passage metric of sampled trajectories.

Constraints (include initial sampling constraints and sampling constraints)：

$$\begin{cases} l_0 = l_{init} \\ t_0 = t_{init} \\ l_{lb} \le l_i \le l_{ub} \\ t_{lb} \le t_i \end{cases} \tag{3}$$

After optimizing the sampled points, using decision-loss estimate the quality of the future driving conditions based on the distribution of curvature longitudinal acceleration and lateral acceleration of the sampled points. A decision loss would be incurred in this scenario.

$$\begin{cases} decision \cos t = w_{area} \sum area\_\cos t + w_{lane} \sum lane\_\cos t \\ + Move \cos t + Dynamic \cos t + Path \cos t \\ Move \cos t = w_{move} \dfrac{longitudinal}{longitudinal\_\max} \\ Path \cos t = w_{11} \mu(kappa) + w_{12} \varepsilon(kappa) \\ Dynamic \cos t = w_{21} \mu(lateral\_acceleration) + w_{21} \varepsilon(lateral\_acceleration) \\ + w_{21} \mu(longitudinal\_acceleration) + w_{21} \varepsilon(longitudinal\_acceleration) \end{cases} \tag{4}$$

$\mu(x), \varepsilon(x)$ Take the expectation and variance of the distribution of variable x.

## 1.2 Path optimization

After the optimal convex space combination is obtained, path optimization is carried out, as shown in Figure4.

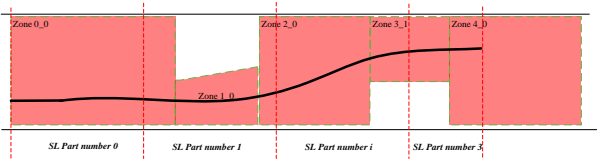

Figure 4 Path optimization under optimal convex combination

According to the expected path length and the length of convex combination, it is divided into multiple SL blocks at a certain interval, and the number is defined as $i$, total number defined as $n$. Set $l$ as a function of $s$ in the SL($i$) block [3]:

$$l = a_0^i + a_0^i s + a_0^i s^2 + a_0^i s^3 + a_0^i s^4 + a_0^i s^5 \quad (5)$$

The cost function as:

$$COST = w_0 \sum_{i=0}^{n} \int_{s_0^i}^{s_1^i} \left(l(s) - l_{target}\right)^2 ds + w_1 \sum_{i=0}^{n} \int_{s_0^i}^{s_1^i} \dot{l}(s)^2 ds + w_2 \sum_{i=0}^{n} \int_{s_0^i}^{s_1^i} \ddot{l}(s)^2 ds + w_3 \sum_{i=0}^{n} \int_{s_0^i}^{s_1^i} \dddot{l}(s)^2 ds + w_{obs} \sum_{i=0}^{n} \int_{s_0^i}^{s_1^i} d(s)^2 ds \quad (6)$$

$\sum_{i=0}^{n} \int_{s_0^i}^{s_1^i} \left(l(s) - l_{target}\right)^2 ds$ defined as the integral of the square of the difference between $l$ and the median line of the target lane. The cost can ensure that the vehicle runs in the target lane.

$\sum_{i=0}^{n} \int_{s_0^i}^{s_1^i} \dot{l}(s)^2 ds$ defined as Integral of $l$ derivative squared. The cost is coordinated with other cost make sure the smoothness of the path when the vehicle takeover obstacles.

$\sum_{i=0}^{n} \int_{s_0^i}^{s_1^i} \ddot{l}(s)^2 ds$、$\sum_{i=0}^{n} \int_{s_0^i}^{s_1^i} \dddot{l}(s)^2 ds$ both of them make sure the smooth of the path.

The link constraints between adjacent SL blocks should be met as shown in Equation (7). Link constraints ensure continuity of paths across SL blocks.

$$\begin{cases} l(s_i^0) = l(s_{i-1}^m) \\ \dot{l}(s_i^0) = \dot{l}(s_{i-1}^m) \\ \ddot{l}(s_i^0) = \ddot{l}(s_{i-1}^m) \end{cases} \quad (7)$$

The starting point of the path must meet the starting point constraint shown in Equation (8), where ***init_sl*** represents the starting point of the path, and the starting point is related to the path of the last frame.

Path function must satisfy position constraint, derivative constraint and second derivative constraint.

$$\begin{cases} l_{min}(s_k) + 0.5w \le l(s_k) + \beta \dot{l}(s_k) \le l_{max}(s_k) - 0.5w \\ \dot{l}_{min}(s_k) \le \dot{l}(s_k) \le \dot{l}_{max}(s_k) \\ \ddot{l}_{min}(s_k) \le \ddot{l}(s_k) \le \ddot{l}_{max}(s_k) \end{cases} \quad (8)$$

$l_{min}(s_k)$、$l_{max}(s_k)$ represents the upper and lower limits of convex space $l$ at $s_k$. $w$ represents the vehicle geometry model width. $\beta$ linearize related parameters, associated with $\dot{l}(s_k)$ and vehicle length, in this paper $\beta$ is 0.2.

At this point, the path optimization problem is transformed into the standard quadratic form [5] of type (9). The number of elements to be optimized is *6n*, and the number of constraints is related to the distance length of convex space.

$$\begin{cases} \min\left(\frac{1}{2}x^T Hx + x^T g\right) \\ s.t. \begin{cases} lbA \le Ax \le ubA \\ lb \le x \le ub \end{cases} \end{cases} \quad (9)$$

### 1.3 Speed optimization

After the optimized path and the meeting time of obstacle vehicles are obtained in the process of an iteration, the obstacle vehicles are projected according to the path and the ST figure [3] is obtained, as shown in Figure 5.

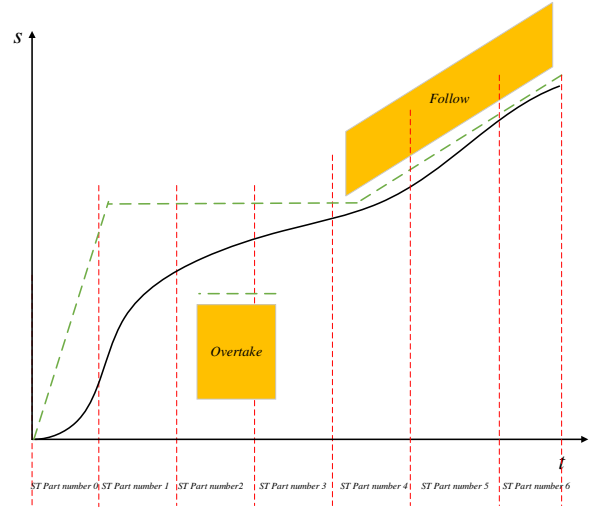

Figure 5 Obstacle decision under ST diagram

According to the meeting time and path, the decision set of obstacle vehicle {yield, follow, overtake} is obtained. The convex space under ST dimension is obtained in the same way as path planning. According to the total planning time, the convex space is divided into multiple ST blocks at a certain time interval, and the number is defined as $i$, total number is defined as $n$.

Setting $s$ as a function of $t$ in the ST block [3]:

$$s = a_0^i + a_0^i t + a_0^i t^2 + a_0^i t^3 + a_0^i t^4 + a_0^i t^5 \quad (10)$$

Cost function as:

$$COST = w_0 \sum_{i=0}^{n} \int_{t_0^i}^{t_1^i} \left(s(t) - s_{up}\right)^2 dt + w_1 \sum_{i=0}^{n} \int_{t_0^i}^{t_1^i} \left(\dot{s}(t) - \dot{s}_{expect}(t)\right)^2 dt + w_2 \sum_{i=0}^{n} \int_{t_0^i}^{t_1^i} \ddot{s}(t)^2 dt + w_3 \sum_{i=0}^{n} \int_{t_0^i}^{t_1^i} \dddot{s}(t)^2 dt \quad (11)$$

$\sum_{i=0}^{n} \int_{t_0^i}^{t_1^i} \left(s(t) - s_{up}\right)^2 dt$ make sure that the vehicle is operating at the speed closest to the road speed limit under constraints.

$\sum_{i=0}^{n} \int_{t_0^i}^{t_1^i} \left(\dot{s}(t) - \dot{s}_{expect}(t)\right)^2 dt$ make sure that the vehicle is as close to lane speed as possible under constraints.

$\sum_{i=0}^{n} \int_{t_0^i}^{t_1^i} \ddot{s}(t)^2 dt$ 、 $\sum_{i=0}^{n} \int_{t_0^i}^{t_1^i} \dddot{s}(t)^2 dt$ make sure vehicle comfort.

The link constraint between adjacent ST blocks should be met as shown in Equation (12). Link constraints ensure continuity of paths across ST blocks.

$$\begin{cases} s(t_i^0) = s(t_{i-1}^m) \\ \dot{s}(t_i^0) = \dot{s}(t_{i-1}^m) \\ \ddot{s}(t_i^0) = \ddot{s}(t_{i-1}^m) \end{cases} \quad (12)$$

The starting point of ST curve meets the starting point constraint as shown in Equation (13), where **init_st** represents the starting point, which is related to the last planned speed.

$$\begin{cases} s(t_0^0) = init\_st.s \\ \dot{s}(t_0^0) = init\_st.velocity \\ \ddot{s}(t_0^0) = init\_st.acc \end{cases} \quad (13)$$

The velocity must satisfy the dynamic constraint, as shown in Equation (14).

$$\begin{cases} s_{k-1} + \dot{s}_{k-1}\Delta t + \frac{1}{2}acc_{min}\Delta t^2 \leq s_k \leq s_{k-1} + \dot{s}_{k-1}\Delta t + \frac{1}{2}acc_{max}\Delta t^2 \\ \dot{s}_{k-1} + acc_{min}\Delta t \leq \dot{s}_k \leq \dot{s}_{k-1} + acc_{max}\Delta t \end{cases} \quad (14)$$

According to the obstacle decision information, ST curve must meet the obstacle related constraints, as shown in Equations (15) ~ (17).

Overtake obstacle:

$$s.t. \begin{cases} s(object)_k^{max} \leq s_k \\ \dot{s}(object)_k \leq \dot{s}_k \end{cases} (t_1 \leq t \leq t_2) \quad (15)$$

Yield obstacle:

$$s.t. \begin{cases} s(object)_k^{min} \geq s_k \\ 0 \geq \dot{s}_k \end{cases} (t_1 \leq t \leq t_2) \quad (16)$$

Follow obstacle:

$$s.t. \begin{cases} s(object)_k^{min} \geq s_k \\ \dot{s}(object)_k \geq \dot{s}_k \end{cases} (t_1 \leq t \leq t_2) \quad (17)$$

$t_1$ stands for the meeting start time and $t_2$ stands for the meeting end time. $s(object)_k^{min}$ represents the minimum $s$ value of the obstacle in the ST diagram during the meeting time, $s(object)_k^{max}$ represents the maximum $s$ value of the obstacle in the ST diagram during the meeting time.

At this point, the speed optimization problem is transformed into the standard quadratic form as shown in Equation (9).

## 2 Iteration

According to the last iteration solution, calculate the **init point** of SL path and ST curve.

Step zero, according to the road speed limit and obstacle prediction information, calculate the meeting time and location of obstacle.

Step $i$, according to the optimization solution and prediction module in step $i$-1, the obstacle prediction information is output, and the time and location of obstacle meeting are calculated one by one.

……

When there is no collision between the vehicle and obstacles, and the error with the last iteration result is less than the limit value, stop iteration.

## Test

There are dynamic and static obstacles in the test lane, and the perception prediction module output normally. The test results show that the planning algorithm can generate smooth path and smooth speed, and the vehicle runs smoothly.

Table1　Environment

| environment | Half-enclosed Road |
| --- | --- |
| Number static obstacle | 1~10 |
| Number dynamic obstacle | 1~5 |
| Path length | 100m |
| Time of speed plan | 7s |

Table2 Test data

| Content | Result |
| --- | --- |
| Average iteration time | 3 |
| Computing time | 20~50ms |

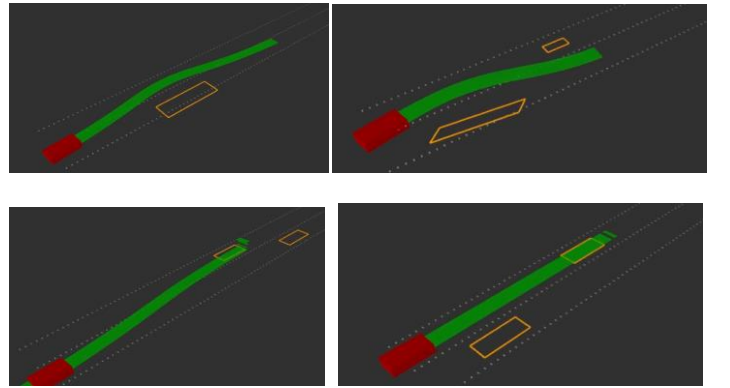

Figure 5 Test display pictures